\pdfoutput=1

\documentclass[11pt]{article}

\usepackage{fontawesome5}

\usepackage[final]{acl}
\usepackage{booktabs} 
\usepackage{multirow} 

\usepackage{times}
\usepackage{latexsym}
\usepackage{array} 
\usepackage{caption} 
\usepackage[T1]{fontenc}

\usepackage[utf8]{inputenc}

\usepackage{microtype}

\usepackage{inconsolata}

\usepackage{graphicx}

\title{Multilingual Encoder Knows more than You Realize:\\Shared Weights Pretraining for Extremely Low-Resource Languages}

\author{%
  Zeli Su\textsuperscript{1,2}\thanks{Equal contribution.}   \
  Ziyin Zhang\textsuperscript{3$*$}  \
  Guixian Xu\textsuperscript{1,2}\thanks{Corresponding author.}  \
  Jianing Liu\textsuperscript{2} \AND
  XU Han\textsuperscript{1,2} \
  Ting Zhang\textsuperscript{1,2} \
  Yushuang Dong\textsuperscript{1,2}\
  \vspace{6pt}\\
  \textsuperscript{1}Key Laboratory of Ethnic Language Intelligent Analysis and Security Governance of MOE \\
  \textsuperscript{2}Minzu University of China \
  \textsuperscript{3}Shanghai Jiao Tong University \\
  \texttt{\{rickamorty,guixian\_xu,hanxu,jianing\_liu,yushuangdong\}@muc.edu.cn}\\
  \texttt{daenerystargaryen@sjtu.edu.cn} ~~~~\
  \texttt{tozhangting@126.com}
}

\begin{document}
\maketitle
\begin{abstract}
While multilingual language models like XLM-R have advanced multilingualism in NLP, they still perform poorly in extremely low-resource languages. This situation is exacerbated by the fact that modern LLMs such as LLaMA and Qwen support far fewer languages than XLM-R, making text generation models non-existent for many languages in the world. To tackle this challenge, we propose a novel framework for adapting multilingual encoders to text generation in extremely low-resource languages. By reusing the weights between the encoder and the decoder, our framework allows the model to leverage the learned semantic space of the encoder, enabling efficient learning and effective generalization in low-resource languages. Applying this framework to four Chinese minority languages, we present XLM-SWCM, and demonstrate its superior performance on various downstream tasks even when compared with much larger models.

\hspace{-10pt}\faGithub {\small~\url{https://github.com/asd765973346/xlm-swcm}}

\hspace{-10pt}\includegraphics[width=1em,height=1em]{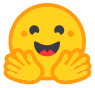}\hspace{-3pt} {\small~\url{https://huggingface.co/KEVVVV/xlm-swcm}}

\end{abstract}


\section{Introduction}\label{sec:intro}

\begin{figure}
    \centering
    \includegraphics[width=1\linewidth]{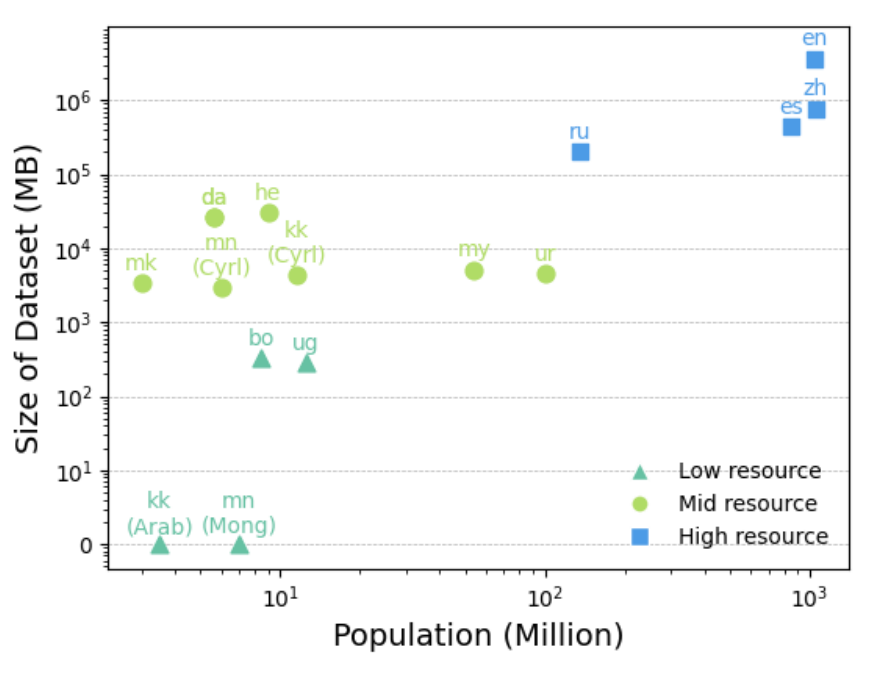}
    \caption{The relationship between population size and dataset size in OSCAR (y-axis, in MB) for various high-, middle-, and low-resource languages. \textbf{Kazakh (kk), Mongolian (mn), Tibetan (bo), and Uyghur (ug)} represent the languages we studied in this work.}
    \label{fig:1_plot}
\end{figure}

In recent years, with the development of multilingual pretrained models such as XLM-R~\citep{xlm-r}, mBART~\citep{mbart}, and mT5~\citep{mt5}, language models have achieved significant progress in multilingual tasks, especially for high-resource languages. However, low-resource languages like Tibetan, Uyghur, Kazakh, and Mongolian—spoken by millions of people in China—remain critically underserved. Among these languages, Tibetan has over 10 million speakers, Uyghur over 11 million, Kazakh approximately 3 million, and Mongolian around 7 million, yet their representation in existing multilingual corpora is vastly inadequate. As illustrated in Figure~\ref{fig:1_plot}, there is a significant disparity between the population sizes of these languages and the amount of available data in popular multilingual corpora such as OSCAR~\citep{OSCAR}. The situation is especially dire for Kazakh and Mongolian, with virtually zero usable data, hindering their inclusion in mainstream multilingual models.

Despite claims of multilingual support for hundreds of languages, models like mBART and mT5 are not trained on Chinese minority languages. In comparison, more advanced multiglingual large language models such as LLaMA~\citep{llama} and Qwen~\citep{qwen2} support even fewer languages. 

\begin{figure*}
    \centering
    \includegraphics[width=\textwidth]{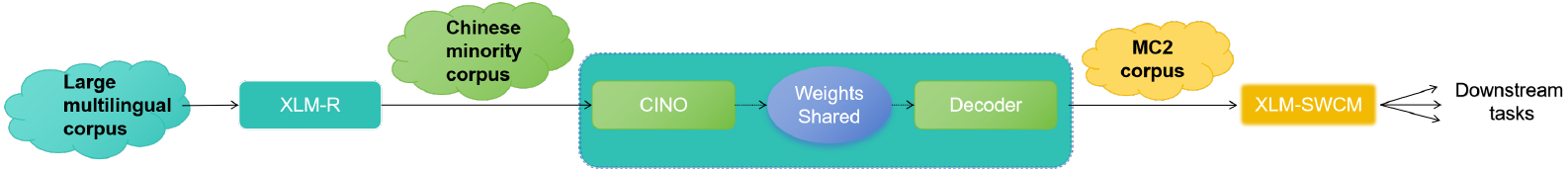} 
    \caption{An overview of the shared weight framework for efficiently adapting multilingual encoders to text generation in low-resource languages.}
    \label{fig:overview-model}
\end{figure*}

This gap underscores the need for targeted solutions to address the challenges of text generation in extremely low-resource languages. To tackle this challenge, we propose a novel framework for efficiently extending a multilingual encoder into an encoder-decoder architecture. To address the scarce training data in low-resource languages, we introduce a weight-sharing mechanism between the encoder and the decoder by interleaving weights transferred from the encoder with randomly initialized ones, allowing for efficient adaptation to text generation in low-resource settings.


Extensive experiments on the aforementioned four Chinese minority languages demonstrate the convincing advantages of our proposed method, with both faster convergence, better generalization, and strong cross-lingual transfer capabilities. Our model, \textbf{XLM-SWCM} (XLM-Shared Weight for Chinese Minorities), outperforms an mBART baseline by up to 199\% on text summarization, 108\% on reading comprehension, and also bests the much larger MC2-LLaMA 13B~\citep{mc2} in cross-lingual transfer settings.

In summary, the main contributions of this paper are:

1) a weight-sharing framework for efficiently adapting multilingual encoders to text generation in low-resource languages;

2) a model XLM-SWCM trained with this method for multiple Chinese minority languages;

3) extensive experiments showcasing the superior performance of XLM-SWCM compared with similar-sized baselines and much larger LLMs, confirming the feasibility of our framework.

Our code and models will be released upon publication.

\section{Related Works}\label{sec:related-work}
\subsection{Multilingual Corpus}
The evolution of multilingual large language models (LLMs) has been enabled by the release of extensive multilingual corpora such as CC100, mC4, OSCAR, CulturaX, and Madlad-400 \cite{CC100,mc4,OSCAR,culturax,Madlad}. While these resources cover a selection of low-resource languages to some extend, there remains a recognized gap in the representation for China's minority languages, primarily due to significant differences in writing systems.
   
China’s minority languages often use different writing systems from the same language family used elsewhere in the world. For example, Uyghur is primarily written in the Arabic script (UEY—Uyghurche Ereb Yëziqi) in China, with the Latin script (ULY—Uyghurche Latin Yëziqi) used as a supplementary form. In contrast, Uyghur in Russia and Central Asia is written in the Cyrillic script (USY—Uyghurche Shilir Yëziqi). When collecting data for minority languages, the aforementioned multilingual corpora either do not distinguish between such different writing systems, or only contain data from one system, as shown in Figure~\ref{fig:1_plot}.
   
Recently, the release of the Multilingual Corpus of Minority Languages in China (MC2,~\citealp{mc2}) breaks the gap in the availability of Chinese minority language pretraining corpora, covering four underrepresented languages: Tibetan, Uyghur, Kazakh, and Mongolian. This dataset is used as the primary pretraining corpus in our work.

\subsection{Development of Multilingual Language Models}
In the past few years, multilingual variants of pretrained language models have been proposed in the NLP community, such as mBART~\citep{mbart} and mT5~\citep{mt5}, supporting up to 100 languages and demonstrating powerful cross-lingual transfer capabilities. More recently, the emergence of large language models (LLMs) has revolutionized multilingual natural language processing. Models like PaLM~\citep{palm} and BLOOM~\citep{bloom} have made significant strides in multilingual capabilities, while the LLaMA family~\citep{llama} and its multilingual variants have democratized access to multilingual LLMs. Some specialized models represented by XGLM and NLLB~\citep{xglm,nllb} have focused on expanding language coverage and improving cross-lingual transfer capabilities across hundreds of low-resource languages. However, few of these models support Chinese minority languages.


\subsection{NLP for Minority Languages in China}
To enhance the accessibility of minority languages in China, prior studies have primarily focused on curating annotated datasets for various NLP tasks. These efforts have mainly concentrated on three key task categories: text classification \cite{bo-classfy,bo-classfy-3}, question answering \cite{boqa}, and machine translation \cite{bomt}. Prominent models specifically trained for these languages include CINO~\citep{cino}, MiLMo~\citep{milmo}, and TiBert~\citep{tibert}. However, despite such progress, none of these models have released their pre-training corpora, and there is still a notable gap in the availability of models capable of text generation in these languages.

\section{Method}\label{sec:method}

\begin{figure}
    \centering
    \includegraphics[width=1\linewidth]{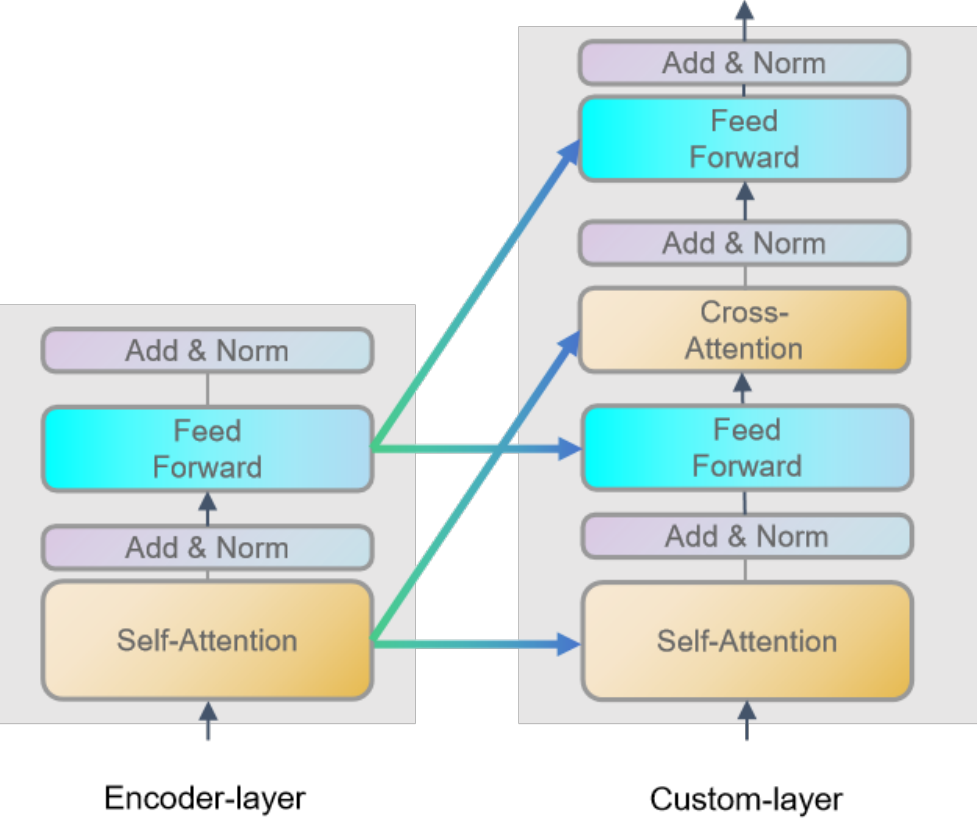}
    \caption{The weight initialization schemes for the CustomDecoderLayer. The colored arrows indicate the initialization of weights between the different components.}
    \label{fig:decoder-initialization}
\end{figure}

\subsection{Adapting Encoders to Text Generation}

\subsubsection{Framework Overview}
In this section, we introduce the Shared Weights Framework, which leverages shared weights between the encoder and decoder for efficiently adapting multilingual encoders to text generation in low-resource languages. 

The overall pipeline is visually summarized in Figure~\ref{fig:overview-model}. Starting from CINO~\citep{cino}, a continual-pretrained version of XLM-R for Chinese minority languages, we copy its weight to initialize the decoder layers for knowledge transfer, and tie some of the weights between encoder and dedocer to enable efficient training. This model, which we name XLM-SWCM, is pretrained on the MC2 corpus and then applied to downstream tasks, including both single-language finetuning and cross-lingual transfer.

\subsubsection{Model Architecture}
Like the vanilla Transformer, the proposed model has two main components:

\paragraph{Encoder:} a pre-trained encoder-only model, specifically CINO, a variant of XLM-R enhanced for Chinese minority languages.

\paragraph{Decoder:} a transformer decoder stack with a specialized weight transfer mechanism. To balance the knowledge acquired during the encoder's large-scale pretraining and new knowledge required for downstream generation tasks, we introduce two types of decoder layers: NormalDeocderLayer and CustomDecoderLayer, both maintaining the same hidden dimension, intermediate size, and number of attention heads as the encoder.
    
\textbf{NormalDecoderLayer}: A standard transformer decoder layer with randomly initialized weights. It follows a conventional architecture with sequential self-attention, cross-attention, and feed-forward network. These layers enable the model to learn generation-specific features from scratch, complementing the knowledge transfered from the encoder.

\textbf{CustomDecoderLayer}: A modified transformer decoder layer that inherits pre-trained weights from the encoder. It features an enhanced structure with two strategically positioned feed-forward networks: FFN1 between self-attention and cross-attention, and FFN2 following cross-attention, each with its own layer normalization and residual connection, as shown in Figure~\ref{fig:decoder-initialization}. CustomDecoderLayer inherits all its weights from the pre-trained encoder to reuse learned representations. 
 

\subsubsection{Weight Sharing Mechanism}
    
In our framework, the pre-trained encoder consists of only self-attention and feed-forward blocks, while the decoder layers require both self-attention and cross-attention mechanisms for effective generation. Thus, special schemes are designed to initialize and reuse the weights, as shown in Figure~\ref{fig:decoder-initialization}.

For weight initialization of CustomDecoderLayers, weights of both self-attention and cross-attention in the decoder are initialized from the encoder's self-attention blocks. Similarly, weights of both two FFN blocks in a decoder layer are initialized from the FFN block in the corresponding encoder layer. This mechanism reduces the effective number of parameters to be learned, accelerating convergence and enabling effective transfer of linguistic knowledge from the pre-trained encoder while maintaining model stability.



A key architectural decision in our framework is the insertion pattern of these layers. After every $X$ CustomDecoderLayers, we insert one NormalDecoderLayer, so that an encoder with $n$ layers would correspond to a decoder with $n + \lfloor n / X\rfloor$ layers. The value of $X$ significantly impacts the model’s generalization capabilities, and its optimal value varies across different model scales. Through extensive experimentation, we find that $X=3$ yields the best performance, and a detailed analysis of how this choice affects the model's performance is discussed in Section~\ref{sec:5.3.2}.

\subsection{Pretraining}
\subsubsection{Pretraining Tasks}
We adopte a multi-task training approach for pretraining. The primary task involves self-supervised learning using mBART's \textbf{denoising auto-encoding (DAE)} strategy. This strategy helps with the model's transition from the encoder's word-level cloze tasks to sequence generation tasks by predicting the masked portions of the input sequence with a decoder.

Additionally, we incorporate \textbf{machine translation} as an auxiliary objective, particularly focusing on translation between Mandarin Chinese and various Chinese minority languages. Specifically, the training data includes bidirectional translation pairs between Mandarin Chinese and the minority languages. This auxiliary objective improves the model's cross-lingual transfer capability, thereby enhancing the model's performance in various low-resource language processing tasks.

\subsubsection{Training Data}
\textbf{THUCNews}~\cite{thucnews} is a Chinese news dataset, derived from historical data from the Sina News RSS feed between 2005 and 2011 and containing approximately 740,000 news articles. From this dataset, we extracted a subset of Simplified Chinese news articles.

\textbf{MC2}~\cite{mc2} provides multilingual data for several Chinese minority languages, including Tibetan, Uyghur, Kazakh, and Mongolian. The specific data volumes are described in detail in Appendix \ref{sec:Dataset Detail}. Together with THUCNews, these monolingual datasets serve as training data for the DAE task.

\textbf{Machine Translation Data }were constructed by leveraging Google Translate to create bidirectional translation pairs between Chinese and four minority languages (Tibetan, Uyghur, Kazakh, and Mongolian). Due to the scarcity of parallel corpora for these low-resource languages, we adopted different strategies: for Uyghur, Kazakh, and Mongolian, we manually aggregated fragmented datasets and converted existing English-minority language pairs to Chinese-minority pairs via reliable English-Chinese translation systems. For Tibetan, we sampled from the relatively abundant TibetanSFT Corpus. \footnote{\href{https://huggingface.co/datasets/shajiu/ParallelCorpusSFT}{https://huggingface.co/datasets/shajiu/ParallelCorpusSFT}}All translations were verified by native speakers to ensure quality, with 2,000 sentence pairs selected per language to create balanced supplementary training data across all four minority languages.

Combining these three corpora, the integrated dataset allows the model to effectively handle both high-resource and low-resource languages, improving its cross-lingual transfer and multilingual capabilities.
\section{Experiments}\label{sec4:experiments}
\subsection{Pretraining}

\paragraph{Training Configuration}
The models are trained for 8 epochs with a peak learning rate of 1e-4, AdamW~\citep{2017AdamW} optimizer, global batch size 600, and a linear learning rate scheduler with a warmup proportion of 0.1. The maximum sequence length is set to 256 tokens, and mixed-precision is enabled to optimize memory usage and training efficiency. To ensure training stability, the norms of gradients are clipped to 1.0. The models are trained on two NVIDIA A800 GPUs, each with 80GB of memory, and the training process takes 92 hours.

\paragraph{Balanced Sampling Strategy}
To address the inherent data imbalance across different languages, we implemente a balanced sampling strategy similar to XLM-R. The sampling probability for each language is calculated as

\begin{equation}
    p_i = \frac{q_i^\alpha}{\sum_j q_j^\alpha},
\end{equation}

\noindent where $q_i$ represents the original proportion of language $i$ in the dataset, and $\alpha$ (set to 0.3) is a smoothing parameter that balances between uniform sampling and size-proportional sampling. This approach ensures that low-resource languages receive adequate representation in the training process while maintaining the influence of larger datasets.

\paragraph{Model Adaptations}
We extende the model's vocabulary with special language tokens (<bo>, <kk>, <mn>, <ug>, <zh>) to handle our target languages (Tibetan, Kazakh, Mongolian, Uyghur, and Chinese). These language identifiers are directly added after the bos token <s> in the model inputs. This modification ensures that the model can effectively process and distinguish between different languages during both pre-training and downstream task finetuning. The same approach is consistently applied in all subsequent experiments.


Based on the aforementioned settings, we trained a new seq2seq model - XLM-SWCM, utilizing CINO-base-v2 as the encoder, with 457 million parameters. The detailed architectural configuration is provided in Appendix \ref{sec:Training Detail}.

\subsection{Downstream Tasks}

\subsubsection{Experiment Setting}\label{sec:experiments-downstream-setting}

To evaluate the capabilities of XLM-SWCM, we conduct fine-tuning experiments on three downstream tasks in both low-resource and high-resource languages: Text Summarization, Machine Reading Comprehension (MRC), and Machine Translation. These tasks are chosen to cover diverse areas of text generation in NLP.

\paragraph{Single-Language Fine-tuning} Due to the scarcity of labeled data for low-resource languages, we focus primarily on Tibetan for single-language fine-tuning, which has several publicly available datasets:

- \textbf{Text Summarization}: For this task, we utilize the Ti-Sum dataset~\cite{ti-sum} with 20,000 pairs of titles and articles.

- \textbf{MRC}: We mainly use the TibetanQA dataset~\cite{tibetanqa} for this task, which claims to contain 20K examples. However, only 2K examples are publicly available. Thus we enrich it by integrating 5K examples from the TibetanSFT Corpus and 3K examples translated from a Chinese MRC dataset \cite{chinese-mrc} using Google Translate. This approach enables us to create a comprehensive dataset consisting of 10K examples.

- \textbf{Machine Translation}: For Machine Translation, we also use the TibetanSFT Corpus, which is cleaned to generate 50,000 parallel Chinese-Tibetan sentence pairs.

\paragraph{Cross-lingual Transfer} In addition to single-language fine-tuning, we also conduct cross-lingual transfer experiments to test XLM-SWCM’s ability to generalize across multiple low-resource languages. This experiment aims to assess the model's performance in Tibetan, Uyghur, Mongolian, and Kazakh after being fine-tuned on a high-resource language (Simplified Chinese) and a very small number of samples in the target languages.

- \textbf{Text Summarization}: For Mandarin Chinese, we use the publicly available LCSTS dataset~\citep{lcsts}, which contains 100K samples scraped from various Chinese portals. For the four minority languages, approximately 3K cleaned samples per language are scraped from language-specific news portals, using the news titles as their summarization.
    
- \textbf{MRC}: For Chinese, we employ the CMRC 2018 dataset~\citep{chinese-mrc}, which consists of 10K samples. For Tibetan, we use 500 samples extracted from the publicly available TibetanQA dataset. For the other three minority languages (Uyghur, Mongolian, Kazakh), we utilize machine translation tools to translate and clean MRC data, ultimately selecting 500 samples per language.

\paragraph{Baseline Models}
We employ two baseline models to ensure broad coverage and robust performance in handling Chinese minority languages. Both models are continually pretrained on the MC2 dataset. The first, \textit{MC2-LLaMA-13B}, is based on LLaMA2-Chinese from previous work.\cite{mc2} The second, \textit{mBART-CM}, is adapted from mBART-cc25 with an expanded vocabulary to include tokens specific to our minority languages.

\paragraph{Training settings}
Both XLM-SWCM and mBART-CM are sequence-to-sequence models that are fine-tuned using standard training configurations. Each of these models is trained for 50 epochs with a batch size of 200 samples to ensure comprehensive learning and optimal performance. MC2-LLaMA-13B model is trained using LoRA~\citep{lora} with a rank of 8 for 3 epochs.

\begin{table*}[ht]
  \centering
  \begin{tabular}{c c ccc ccc ccc}
    \toprule
    \multirow{2}{*}{\textbf{Model}} & \multirow{2}{*}{\textbf{Size}}
    & \multicolumn{3}{c}{\textbf{Sum}}
    & \multicolumn{3}{c}{\textbf{MRC}}
    & \multicolumn{3}{c}{\textbf{MT}} \\[0.5ex]
    \cmidrule(lr){3-5}\cmidrule(lr){6-8}\cmidrule(lr){9-11}
    & 
    & \textbf{F} & \textbf{P} & \textbf{R}
    & \textbf{F} & \textbf{P} & \textbf{R}
    & \textbf{F} & \textbf{P} & \textbf{R}\\[0.5ex]
    \midrule
    MC2-LLaMA-13B & 13B
    & 16.1 & 12.3 & 15.5
    & 13.2 & 11.7 & 13.1
    & 15.1 & 12.2 & 16.8 \\[0.5ex]
    mBART-CM & 611M
    & 8.6 & 11.2 & 15.2
    & 7.9  & 6.1 & 5.6
    & 11.5 & 7.3 & 9.3 \\[0.5ex]
    XLM-SWCM (ours) & 492M
    & \textbf{25.7} & \textbf{29.1} & \textbf{24.2}
    & \textbf{16.4} & \textbf{29.5} & \textbf{16.2}
    & \textbf{24.5} & \textbf{26.3} & \textbf{24.3} \\[0.5ex]
    \bottomrule
  \end{tabular}
  \caption{\label{single}
    Performance metrics of the baseline models, evaluated using three ROUGE-L sub metrics: 
    \textbf{F} (F1-score), \textbf{P} (precision), 
    and \textbf{R} (recall). Size refers to the number of parameters in each model.
  }
\end{table*}

\begin{table*}[ht]
    \centering
    \begin{tabular}{l cc cc cc cc cc}
        \toprule
        \multirow{2}{*}{\textbf{Model}} 
        & \multicolumn{2}{c}{\textbf{Zh}} 
        & \multicolumn{2}{c}{\textbf{Bo}}
        & \multicolumn{2}{c}{\textbf{Ug}} 
        & \multicolumn{2}{c}{\textbf{Mn}} 
        & \multicolumn{2}{c}{\textbf{Kk}} \\[0.5ex]
        \cmidrule(lr){2-3}\cmidrule(lr){4-5}\cmidrule(lr){6-7}\cmidrule(lr){8-9}\cmidrule(lr){10-11}
        & \textbf{Sum} & \textbf{MRC}
        & \textbf{Sum} & \textbf{MRC}
        & \textbf{Sum} & \textbf{MRC}
        & \textbf{Sum} & \textbf{MRC}
        & \textbf{Sum} & \textbf{MRC} \\[0.5ex]
        \midrule
        MC2-LLaMA-13B    & 47.1 & 43.5 & 9.5 & 6.1 & 3.5 & 2.4 & 3.7 & 2.2 & 2.6 & 3.9 \\[0.5ex]
        MC2-LLaMA-13B*   & \textbf{47.3} & \textbf{44.7} & 13.1 & \textbf{11.5} & 11.7 & 10.1 & 9.7 & \textbf{10.2} & 2.9 & 4.6 \\[0.5ex]
        mBART-CM     & 32.7 & 25.6 & 6.8 & 2.1 & 2.7 & 2.2 & 3.1 & 1.7 & 0.2 & 0.1 \\[0.5ex]
        XLM-SWCM (ours)     & 33.1 & 23.5 & \textbf{17.1} & 11.1 & \textbf{12.5} & \textbf{11.1} & \textbf{13.5} & 7.2 & \textbf{5.6} & \textbf{6.9} \\[0.5ex]
        \bottomrule
    \end{tabular}
    \caption{\label{fewshot}
     Cross-lingual Transfer performance of different models on Text Summarization (Sum) and Machine Reading Comprehension (MRC) tasks, evaluated using ROUGE-L. The best results for each task are highlighted. *~indicates explicitly prompting MC2-LLaMA-13B with the language to be used in the response during evaluation.
    }
\end{table*}

\subsubsection{Experimental Results}

As illustrated in Table~\ref{single}, XLM-SWCM consistently outperforms the baseline models across all three tasks. Despite having fewer parameters, XLM-SWCM demonstrates a substantial margin of superiority over mBART-CM and even surpasses the much larger MC2-LLaMA-13B.

Notably, XLM-SWCM achieves an impressive \textbf{198.8\% improvement in F1-score for Text Summarization} over mBART-CM, along with a significant \textbf{107.6\% F1 improvement in MRC}. These remarkable gains are a direct result of XLM-SWCM's efficient weight sharing framework to maximize the utilization of pre-trained encoder features in resource-constrained scenarios. Even under equivalent seq2seq structures and identical training corpora, XLM-SWCM demonstrates greater efficiency and learning capacity.

In comparison to MC2-LLaMA-13B, which benefits from richer pretraining corpora and larger-scale parameters, XLM-SWCM achieves a \textbf{59\% higher F1-score in Text Summarization}, a \textbf{24.1\% F1 improvement in MRC}, and a \textbf{62.3\% higher F1-score in MT}. These results underscore the effectiveness of XLM-SWCM's shared weight framework in resource-constrained environments, making it a superior choice for tasks involving Chinese minority languages.

Table~\ref{fewshot} highlights the performance of XLM-SWCM and baseline models in cross-lingual transfer settings. For the primary source language (Zh), the baseline models demonstrate better performance, which stems from their larger parameter sizes and more extensive pretraining corpora in Simplified Chinese. However, when it comes to \textbf{generalization to minority languages}, XLM-SWCM showcases exceptional adaptability, significantly outperforming the baseline models. mBART-CM, for instance, struggles to distinguish between languages and often defaults to outputs in the primary language (Zh), even when language-specific labels are present. Similarly, MC2-LLaMA-13B exhibits language-related errors, though its performance improves when explicitly informed of the current language type, as seen with MC2-LLaMA-13B*.

In Text Summarization, XLM-SWCM outperforms all baselines. Specifically, XLM-SWCM achieves significant improvements of \textbf{30.5\%, 6.8\%, and 39.1\%} for Tibetan (Bo), Uyghur (Ug), and Mongolian (Mn) respectively over MC2-LLaMA-13B*, the best-performing baseline. For MRC, XLM-SWCM also demonstrates competitive performance across most languages, being only slightly weaker than MC2-LLaMA-13B* for Tibetan and Mongolian. 

Overall, these experiments indicate that XLM-SWCM can effectively leverage the shared weight mechanism to maximally reuse the semantic space of the pre-trained encoder, demonstrating excellent performance in Chinese minority language applications with limited data and parameter size.

\begin{table}[ht]
  \centering
  \resizebox{\columnwidth}{!}{
  \begin{tabular}{c c c c c}
    \toprule
    \textbf{Removing Module} & \textbf{Sum} & \textbf{MRC} & \textbf{MT} \\[0.5ex]
    \midrule
    None (XLM-SWCM) & \textbf{25.7} & \textbf{16.4} & \textbf{24.5} \\[0.5ex]
    MT & 25.6 & 15.1 & 20.3 \\[0.5ex]
    DAE & 22.4  & 12.2 & 18.7 \\[0.5ex]
    WS & 17.1  & 11.7 & 18.2 \\[0.5ex]
    MT + DAE & 22.5 & 12.3 & 17.7 \\[0.5ex]
    MT + WS  & 17.5 & 11.3 & 18.4 \\[0.5ex]
    DAE + WS & 15.2  & 11.9  & 17.1 \\[0.5ex]
    MT + DAE + WS & 15.9 & 10.8 & 16.5 \\[0.5ex]
    \bottomrule
  \end{tabular}
  }
  \caption{\label{ablation-single}
    Objective ablation results, evaluated using ROUGE-L.
    The experiments involve removing different combinations of training components, such as Machine Translation (MT), DAE (Denoising Auto-Encoding), and Weight Sharing (WS).
  }
\end{table}

\section{Ablation Studies}\label{sec:ablation}
In this section, we present a series of ablation experiments aimed at evaluating the impact of key components in our framework that play essential roles in enhancing the model’s multilingual capabilities and improving its generalization to low-resource languages. We perform ablation experiments on the Tibetan finetuning tasks, maintaining a consistent finetuning setting with Section~\ref{sec:experiments-downstream-setting}.

\subsection{Objective  Ablation}
We first focus on three critical aspects of the model: DAE pretraining, machine translation, and weight initialization by removing each and combinations of them. The results are shown in Table~\ref{ablation-single}. Removing any of the three components is detreimental to performance, specifically:

- Machine Translation (MT): Removing machine translation has a relatively small impact on performance across tasks, as shown by both individual removal (maintaining 25.6 in Sum) and combined removals (MT+DAE vs DAE showing similar scores);
    
- Denoising Auto-Encoding (DAE): The removal of DAE pretraining causes considerable performance drops across all three downstream tasks, and its impact becomes more pronounced in combined removals (DAE+WS), indicating its fundamental importance in establishing the model's basic text generation capabilities.

- Weight Sharing (WS): The removal of weight sharing demonstrates the most significant impact among all modules, showing the largest performance drops in individual removal and maintaining this substantial negative effect across all combined removal scenarios, establishing it as the most crucial component for the model's effectiveness in low-resource settings.

In short, while all three components contribute positively to the model's performance, weight sharing emerges as the most critical component. This finding highlights the importance of weight sharing as a key architectural choice for multilingual models, especially in resource-constrained scenarios.


\subsection{Structure Ablation}
We also perform experiments to evaluate the impact of different structural components in our proposed framework. These experiments aim to understand how the initialization of decoder weights and the insertion of normal layers affect model performance.

\subsubsection{Impact of Weight Initialization}
Firstly, we train a baseline model called \textbf{Cino-Transformer}. Unlike XLM-SWCM, the decoder of this model is randomly initialized, and also matches the number of encoder layers. The model is pretrained using the same DAE and MT tasks as XLM-SWCM but without weight sharing, and then finetuned on downstream tasks in the same setting as XLM-SWCM.

\begin{table}[ht]
  \centering
  \resizebox{\columnwidth}{!}{
  \begin{tabular}{c c c c c}
    \toprule
    \textbf{Model} & \textbf{Sum} & \textbf{MRC} & \textbf{MT} \\[0.5ex]
    \midrule
    Cino-Transformer & 18.9 & 13.5 & 18.7 \\[0.5ex]
    XLM-SWCM (ours) & \textbf{25.7} & \textbf{16.4} & \textbf{24.5} \\[0.5ex]
    \bottomrule
  \end{tabular}
  }
 \caption{\label{ablation-structure}
   Performance metrics of the Ablation of Weight Initialization, evaluated using the ROUGE-L metric. 
}
\end{table}

\begin{table}[ht]
  \centering
  \resizebox{\columnwidth}{!}{
  \begin{tabular}{c c c c c}
    \toprule
    \textbf{Model} & \textbf{Sum} & \textbf{MRC} & \textbf{MT} \\[0.5ex]
    \midrule
    BASE-A & 13.7 & 10.3 & 15.7 \\[0.5ex]
    BASE-B & 16.3 & 14.1 & 21.1 \\[0.5ex]
    XLM-SWCM (ours) & \textbf{25.7} & \textbf{16.4} & \textbf{24.5} \\[0.5ex]
    \bottomrule
  \end{tabular}
  }
\caption{\label{ablation-Normal-Layers}
   Performance metrics of the Ablation of Normal Layers, evaluated using the ROUGE-L metric. 
   \textbf{BASE-A} has fewer layers and does not include any normal layers, while \textbf{BASE-B} maintains the same number of layers as XLM-SWCM but uses weight duplication instead of normal layers. 
}

\end{table}

The results in Table \ref{ablation-structure} demonstrate the effectiveness of our weight initialization scheme. By transferring weights from the encoder to the decoder, XLM-SWCM can be efficiently adapted to text generation with limited training data, outperforming Cino-Transformer on all tasks.

\subsubsection{Impact of Randomly Initialized Layers}

Secondly, we explore the impact of inserting normal layers among the custom layers in the decoder. To assess the effectiveness of this modification, we use two baseline models for comparison:

- \textbf{Baseline A (XLM-SWCM without normal layers)}: This model is identical to XLM-SWCM but without any normal layers inserted into the custom layer architecture. The absence of normal layers leads to a reduced total number of layers in the decoder.

- \textbf{Baseline B (Weight duplication model)}: Instead of inserting normal layers, this model simply copies the weights of the preceding layer to maintain consistency in the number of model parameters. This results in identical weights across consecutive layers, allowing us to isolate the impact of inserting randomly initialized normal layers.

The results in Table~\ref{ablation-Normal-Layers} demonstrate the significant impact of inserting normal layers into the decoder. BASE-A, which has fewer layers, performs the worst across all tasks. BASE-B, which maintain the same number of layers as XLM-SWCM but lacks randomly initialized weights, shows some improvement but still underperforms.

Overall, these findings indicate that randomly initialized normal layers is also crucial for adapting encoders to text generation.

\subsubsection{Impact of Insertion Frequency of Normal Layers}
\label{sec:5.3.2}
Thirdly, we thoroughly investigate the impact of insertion frequency of normal layers in the decoder, and how this interacts with varying dataset sizes. This experiment is designed along two dimensions:

- \textbf{Insertion Frequency of Normal Layers}: we explore values of \( X \) where a normal layer is inserted after every \( X \) custom layers, with \( X \) ranging from 1 to 6. All these models are pretrained in the same setting as XLM-SWCM.

\begin{figure}
    \centering
    \includegraphics[width=1\linewidth]{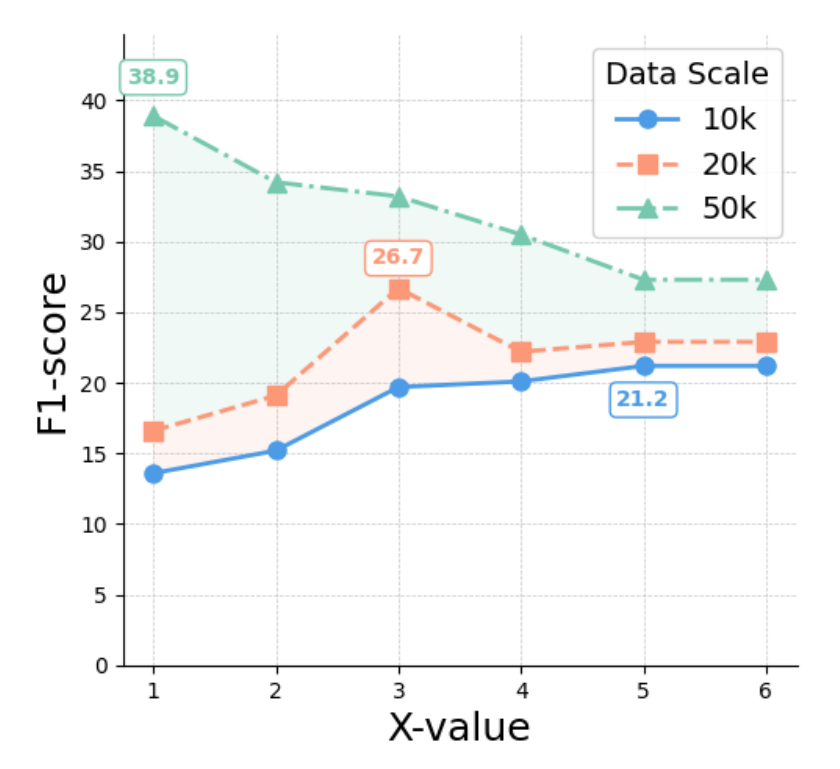}
    \caption{ROUGE-L scores on Tibetan summarization for different X-values (insertion frequency of normal layers). The three lines correspond to different dataset sizes.}
    \label{fig:enter-label}
\end{figure}

- \textbf{Effect of Finetuning Dataset Size}: we evaluate the model’s performance on datasets of varying sizes, including 10K, 20K, and 50K samples. As the existing Ti-SUM dataset only has 20K samples, we supplement it by crawling and cleaning 30K additional news articles from various major Chinese websites. This dimension allows us to examine the interaction between the amount of available data and the frequency of normal layers.

The results are plotted in Figure~\ref{fig:enter-label}:

- For the small dataset (10k), larger $X$ results in better performance, as smaller decoders generalize more effectively when data is limited. In contrast, smaller $X$ (i.e. larger decoders) leads to overfitting.

- For the medium dataset (20k), performance peaks at \( X = 3 \). This indicates that a moderate decoder size strikes a balance between capacity and data availability.

- For the large dataset (50k), smaller $X$ achieve the highest F1-scores, as the larger decoder capacity enables the model to fully exploit the larger dataset.

Overall, these results demonstrate the flexibility of our framework, where the insertion frequency of normal layers can be adjusted based on the task-specific dataset size. Larger $X$ (fewer layers) is better suited for small datasets, while smaller $X$ (more layers) performs best on larger datasets.
\section{Conclusion}\label{sec:conclusion}
In this work, we proposed a novel pretraining framework tailored for low-resource languages, with a particular focus on Chinese minority languages. Our framework leverages a shared weight mechanism between the encoder and decoder, which allows for the efficient adaptation of multilingual encoders to generation tasks without the need to start from scratch. Experimental results demonstrate that our model XLM-SWCM significantly outperforms traditional baselines on various text generation tasks for Tibetan, Uyghur, Kazakh, and Mongolian, which have long been underserved in NLP research. Our approach opens up new possibilities for developing robust models for these extremely low-resource languages, and also provides a promising method for the integration of resources across similar languages.


\section*{Limitations}
Due to the availability of pretrained language models for Chinese minority languages and high-quality corpora, our study focused on only four minority languages. Our single-language finetuning experiments are further constrained to Tibetan given the lack of relevant datasets, limiting the scope of our exploration.

Thus, we hope that future work will put more focus on the development of high-quality datasets in these minority languages and beyond, enabling a more thorough exploration of underrepresented languages in the LLM era. As more data becomes available and the model's capabilities continue to improve, the exploration of these languages will become a key direction for future research.

\section*{Acknowledgements}
This research was supported by the Joint Research Project of Li'an International Education Innovation Pilot Zone, Hainan Province, China (Grant No: 624LALH006).


\bibliography{reference}

\clearpage
\appendix
\onecolumn

\section{Dataset Details}\label{sec:Dataset Detail}

For pretraining of XLM-SWCM and other baseline models, we used a combination of Simplified Chinese data from THUCNews and minority languages from MC2. The breakdown of their distribution is given in Table~\ref{dataset-table}.

\begin{table*}[ht]
  \centering
  \begin{tabular}{>{\centering\arraybackslash}m{3.5cm} >{\centering\arraybackslash}m{2.3cm} >{\centering\arraybackslash}m{2.8cm} >{\centering\arraybackslash}m{3.5cm}}
    \toprule
    \textbf{Language} & \textbf{Data Size} & \textbf{Number of Samples} & \textbf{Tokens} \\[0.5ex]
    \midrule
    Tibetan (bo) & 2.2 GB & 184,045 & 131,531,254 \\[0.5ex]
    Uyghur (ug) & 736 MB & 90,441 & 85,514,016 \\[0.5ex]
    Kazakh (kk) & 937 MB & 57,827 & 92,730,961 \\[0.5ex]
    Mongolian (mn) & 970 MB & 171,847 & 142,370,017 \\[0.5ex]
    Simplified Chinese (zh) & 2.1 GB & 836,075 & 519,615,707 \\[0.5ex]
    \bottomrule
  \end{tabular}
 \caption{\label{dataset-table}
   Statistics of our pretraining dataset.
}
\end{table*}

\section{Training Details}\label{sec:Training Detail}
In addition to the settings presented in the main paper, here we detail other parameters used during pre-training XLM-SWCM for complete reproduction:

\subsection*{Hardware and Software Configuration}
\phantom{.}\vspace{-14pt}

- \textbf{Hardware:} NVIDIA Tesla A800 GPU, 80 GB RAM * 2, Intel i7 CPU.

- \textbf{Software:} Ubuntu 20.04, CUDA 11.7, PyTorch 2.3

\subsection*{Training Configurations}
\phantom{.}\vspace{-14pt}

- \textbf{Total Training Samples:} 1,340,235

- \textbf{Local Batch Size:} 75

- \textbf{Gradient Accumulation Steps:} 4

- \textbf{Global Batch Size:} 600

- \textbf{Epochs:} 8

- \textbf{Total Training Steps:} 17,864

- \textbf{Optimizer:} AdamW with $\beta_1 = 0.9$, $\beta_2 = 0.999$

- \textbf{Learning Rate:} 1e-4

- \textbf{Warm-up:} Linear warm-up for the first  epoch, gradually increasing the learning rate from 1e-5 to 1e-4.

- \textbf{Scheduled Sampling:} In the first epoch, teacher forcing is applied to guide the model. Subsequently, the teacher forcing ratio is gradually decreased in a linear fashion, transitioning to scheduled sampling~\citep{2015scheduled-sampling}.

\end{document}